%% file: main.tex
\documentclass[runningheads]{llncs}

\usepackage{eccv}

\usepackage{eccvabbrv}

\usepackage{graphicx}
\usepackage{booktabs}

\usepackage[accsupp]{axessibility}  % Improves PDF readability for those with disabilities.

\usepackage{amsmath}
\usepackage{amssymb}
\usepackage{bbm}
\usepackage{multirow}
\usepackage{multicol}
\usepackage{stfloats}
\usepackage[normalem]{ulem}
\usepackage{tabularray}
\usepackage{rotating}
\usepackage{tabularx}
\usepackage{pifont}              % for \ding{51} \ding{55} in tables

\graphicspath{{pics/}{secs/}}

\usepackage{hyperref}

\usepackage{orcidlink}

\usepackage[capitalize]{cleveref}

\begin{document}

\title{MV-SAM3D: Adaptive Multi-View Fusion for Layout-Aware 3D Generation} 

\titlerunning{Abbreviated paper title}

\author{Baicheng Li*\inst{1} \and
Dong Wu*\inst{1} \and
Jun Li\inst{2} \and
Shunkai Zhou\inst{1} \and
Lusong Li*\inst{2} \and
Zecui Zeng*\inst{2} \and
Hongbin Zha*\inst{1}}

\authorrunning{B. Li et al.}

\institute{Peking University \and
JD Explore Academy}

\maketitle

\input{secs/0_abstract}

% Teaser figure — placed here to ensure page-1 positioning
\begin{figure*}[t]
    \centering
    \includegraphics[width=\linewidth]{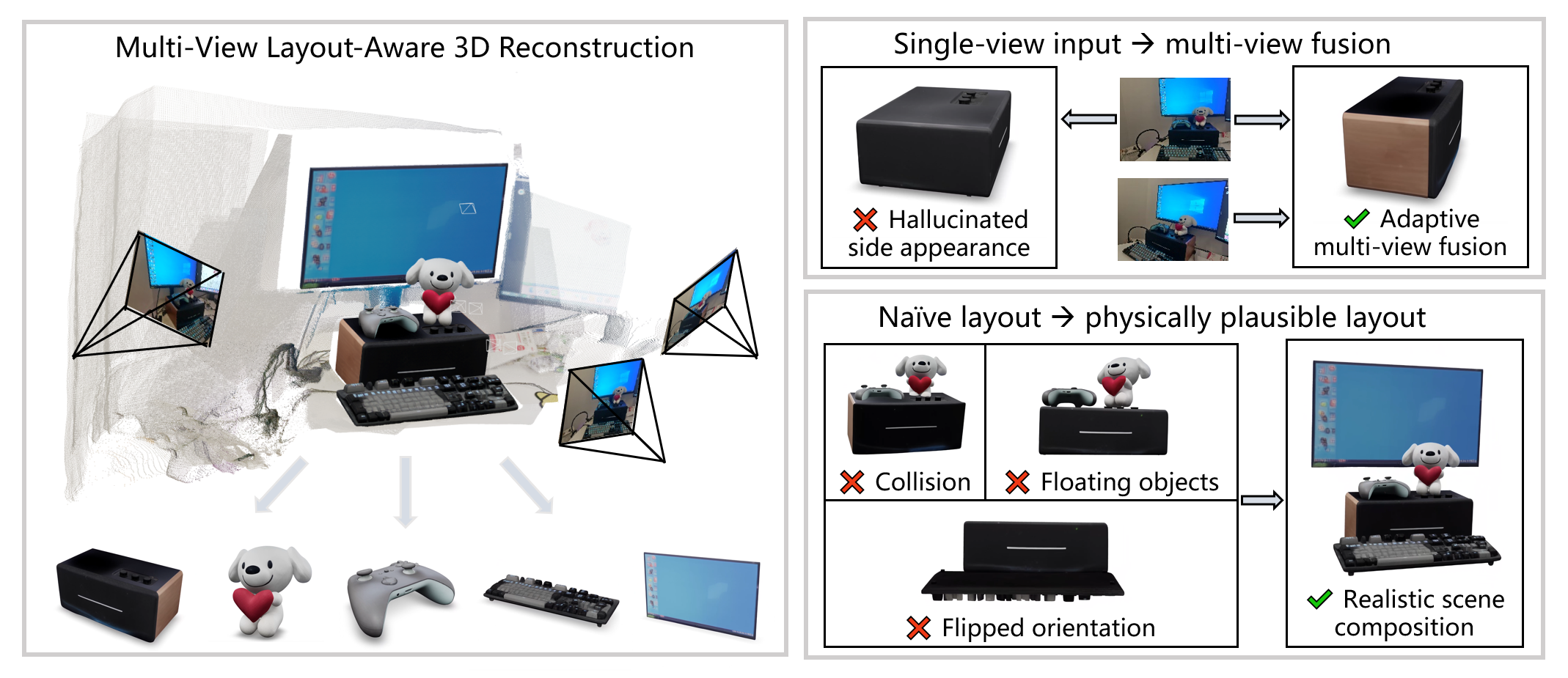}
    \caption{\textbf{MV-SAM3D} enables multi-view, layout-aware 3D generation with physical plausibility. \textbf{Left}: A representative scene-level reconstruction, where each generated 3D object is overlaid onto the scene point cloud. \textbf{Top right}: Single-view generation produces hallucinated side appearance, while our adaptive multi-view fusion yields faithful reconstruction by leveraging complementary observations. \textbf{Bottom right}: Independent pose estimation leads to collisions, floating objects, and incorrect orientations; our physics-aware optimization produces physically plausible object layouts.}
    \label{fig:teaser}
\end{figure*}

\input{secs/1_intro}
\input{secs/2_related_work}
\input{secs/3_method}
\input{secs/4_experiments}

\bibliographystyle{splncs04}
\bibliography{main}
\end{document}

%% file: secs/0_abstract.tex
\begin{abstract}
Recent unified 3D generation models have made remarkable progress in producing high-quality 3D assets from a single image. Notably, layout-aware approaches such as SAM3D can reconstruct multiple objects while preserving their spatial arrangement, opening the door to practical scene-level 3D generation. However, current methods are limited to single-view input and cannot leverage complementary multi-view observations, while independently estimated object poses often lead to physically implausible layouts such as interpenetration and floating artifacts.

We present \textbf{MV-SAM3D}, a training-free framework that extends layout-aware 3D generation with multi-view consistency and physical plausibility. We formulate multi-view fusion as a Multi-Diffusion process in 3D latent space and propose two adaptive weighting strategies --- \emph{attention-entropy weighting} and \emph{visibility weighting} --- that enable confidence-aware fusion, ensuring each viewpoint contributes according to its local observation reliability. For multi-object composition, we introduce physics-aware optimization that injects collision and contact constraints both during and after generation, yielding physically plausible object arrangements. Experiments on standard benchmarks and real-world multi-object scenes demonstrate significant improvements in reconstruction fidelity and layout plausibility, all without any additional training. Code is available at \url{https://github.com/devinli123/MV-SAM3D}.
\end{abstract}

%% file: secs/1_intro.tex
\section{Introduction}
\label{sec:intro}

The generation of 3D content from images has undergone a dramatic transformation in recent years. Driven by large-scale generative models, a new generation of unified 3D reconstruction systems~\cite{trellis,hunyuan3d,xu2024instantmesh} can produce high-fidelity 3D assets from minimal user input, serving as a key enabler for applications from virtual reality to robotics.

Among these advances, SAM3D~\cite{sam3d} introduces \textbf{layout-aware generation}: given a scene image and segmentation masks, it reconstructs each object with high-quality geometry and texture \emph{at its correct position} within the scene. This scene-level reasoning capability distinguishes SAM3D from object-centric alternatives and opens the door to practical multi-object scene reconstruction.

Despite this progress, two challenges remain (see~\cref{fig:teaser}):

\noindent\textbf{Challenge 1: Multi-View Consistency.}
Current layout-aware methods accept only a single image, yet practical scenarios often provide multiple views. Leveraging this information is non-trivial: single-view generation inherently hallucinates unobserved regions, and the model cannot distinguish \emph{observed} content from \emph{imagined} content. When multiple views are naively fused, unreliable hallucinations can overwhelm the reliable observations, degrading rather than improving quality (\cref{fig:teaser}, top right).

\noindent\textbf{Challenge 2: Physical Plausibility in Multi-Object Scenes.}
When generating multi-object scenes, each object's pose is estimated independently without inter-object physical constraints, frequently leading to interpenetration, floating artifacts, and incorrect orientations (\cref{fig:teaser}, bottom right). These physically implausible configurations undermine scene realism and limit downstream applicability.

\noindent\textbf{Our Approach.}
We address both challenges through carefully designed, training-free interventions within the generation pipeline.

For multi-view consistency, we extend SAM3D to accept arbitrary viewpoints by formulating the generation process as a Multi-Diffusion~\cite{bartal2023multidiffusion} in 3D latent space, where velocity predictions conditioned on different views are fused at each generation step. Building on this formulation, we propose two complementary adaptive weighting strategies to further enhance fusion quality: (1)~\emph{attention-entropy weighting}, which leverages the entropy of the model's cross-attention to implicitly estimate observation confidence at each spatial location, and (2)~\emph{geometric visibility weighting}, which explicitly determines whether a 3D point is visible from a given camera based on the reconstructed geometry. Together, these strategies enable confidence-aware fusion where each viewpoint is trusted according to its local reliability, yielding faithful reconstruction across all observed regions.

For physical plausibility, we inject collision and contact constraints directly into the flow matching trajectory during generation, steering it toward plausible configurations. A subsequent refinement stage resolves residual violations on the output meshes.

Based on the above, our contributions are summarized as follows:
\begin{itemize}
    \item We present \textbf{MV-SAM3D}, a training-free framework that extends layout-aware 3D generation from single-view to multi-view input and enhances multi-object scene composition with physical plausibility, without any architectural changes or retraining.
    \item We propose two adaptive weighting strategies --- attention-entropy weighting and visibility weighting --- for confidence-aware multi-view fusion in 3D latent space, ensuring the generated 3D representation faithfully reflects observations from all viewpoints.
    \item We introduce a physics-aware pose optimization pipeline combining generation-time layout injection with post-generation refinement, producing multi-object scenes that respect non-penetration and surface contact constraints.
\end{itemize}

% Teaser figure moved to main.tex (after abstract) for page-1 placement

%% file: secs/2_related_work.tex
\section{Related Work}
\label{sec:related}

\noindent\textbf{3D Generation.}
Feed-forward approaches, including LRM~\cite{hong2024lrm}, InstantMesh~\cite{xu2024instantmesh}, Unique3D~\cite{wu2024unique3d}, TRELLIS~\cite{trellis}, and SPAR3D~\cite{spar3d}, directly regress 3D representations from a single image in a single forward pass, achieving near-real-time performance. More recently, native 3D diffusion models such as Hunyuan3D 2.0~\cite{hunyuan3d}, Direct3D~\cite{direct3d}, TripoSG~\cite{triposg}, and CraftsMan3D~\cite{craftsman3d} generate high-fidelity 3D assets directly in structured latent or mesh spaces, bypassing multi-view intermediate representations. Earlier optimization-based methods built on Score Distillation Sampling~\cite{poole2023dreamfusion,lin2023magic3d} distill knowledge from pretrained 2D diffusion models but remain slow and prone to artifacts. Among these approaches, SAM3D~\cite{sam3d} uniquely introduces \emph{layout-aware} generation, predicting not only object geometry and texture but also each object's spatial arrangement within the scene; Fast-SAM3D~\cite{fastsam3d} further accelerates this via heterogeneity-aware optimization. On the multi-view side, SyncDreamer~\cite{liu2024syncdreamer} and Wonder3D~\cite{long2024wonder3d} synthesize multi-view images from a single input and lift them to 3D, while EscherNet~\cite{kong2024eschernet} enables flexible view-conditioned generation from arbitrary camera poses. However, these multi-view methods focus on object-centric reconstruction and lack scene-level layout awareness. Our work extends layout-aware generation to accept multi-view input, enhancing multi-object compositional generation through more reliable observation coverage and inter-object reasoning.

\noindent\textbf{Multi-View 3D Reconstruction.}
Recovering 3D structure from multiple images has a long history. Classical approaches based on Structure-from-Motion (SfM)~\cite{schoenberger2016sfm} and visual SLAM~\cite{muratal2015orbslam} solve camera poses and sparse/dense geometry through feature matching and bundle adjustment, but require many views and struggle with textureless regions. Learning-based methods have recently made dramatic progress. DUSt3R~\cite{wang2024dust3r} and its extension MASt3R~\cite{leroy2024mast3r} reformulate stereo reconstruction as a regression problem, predicting dense pointmaps from image pairs in a single forward pass. VGGT~\cite{wang2025vggt} and Fast3R~\cite{fast3r} further scale this paradigm to arbitrary numbers of views via feed-forward transformers. Depth Anything 3 (DA3)~\cite{yang2024da3} recovers metrically consistent geometry and camera poses from uncalibrated multi-view images, providing a unified foundation for downstream tasks. In our framework, DA3 replaces the monocular depth estimator (MOGE) used in SAM3D: its metric-scale pointmaps serve as the geometric input for generation, while the estimated camera poses enable explicit visibility computation and multi-object pose optimization. Our multi-view fusion approach shares a common spirit with these reconstruction methods in aggregating observations from multiple viewpoints to reduce reliance on hallucination. However, unlike pure reconstruction, our generative backbone ensures plausible completion for regions that remain unobserved across all views.

\noindent\textbf{Compositional Scene Generation and Reconstruction.}
Generating and reconstructing 3D scenes at the object level has attracted growing attention. Object-compositional reconstruction methods such as ObjectSDF++~\cite{wu2023objectsdfpp} decompose multi-view images into per-object neural implicit surfaces, while PhyRecon~\cite{ni2024phyrecon} further integrates differentiable physics simulation to ensure physically plausible geometry during reconstruction. On the generation side, layout-guided approaches compose scenes by generating objects individually and placing them in a shared coordinate frame: GALA3D~\cite{zhou2024gala3d} optimizes layout-guided Gaussian splatting scenes, GraphDreamer~\cite{gao2024graphdreamer} decomposes complex scene descriptions via scene graphs for node-wise object generation, and ComboVerse~\cite{chen2024comboverse} applies spatially-aware diffusion guidance for multi-component asset creation. Going further, PhyCAGE~\cite{yan2024phycage} introduces physics simulation to resolve inter-part penetrations in compositional 3D generation from a single image. Classifier guidance~\cite{dhariwal2021diffusion} and its training-free variants~\cite{bansal2023universal,yu2023freedom} demonstrate that external objectives can steer diffusion trajectories; PhysDiff~\cite{yuan2023physdiff} applies this principle with physics constraints for human motion. Our work also operates at the object level, generating individual 3D assets together with their layout parameters and composing them into physically plausible scenes through physics-aware pose optimization.

%% file: secs/3_method.tex
\section{Method}
\label{sec:method}

\begin{figure*}[t]
    \centering
    \includegraphics[width=\linewidth]{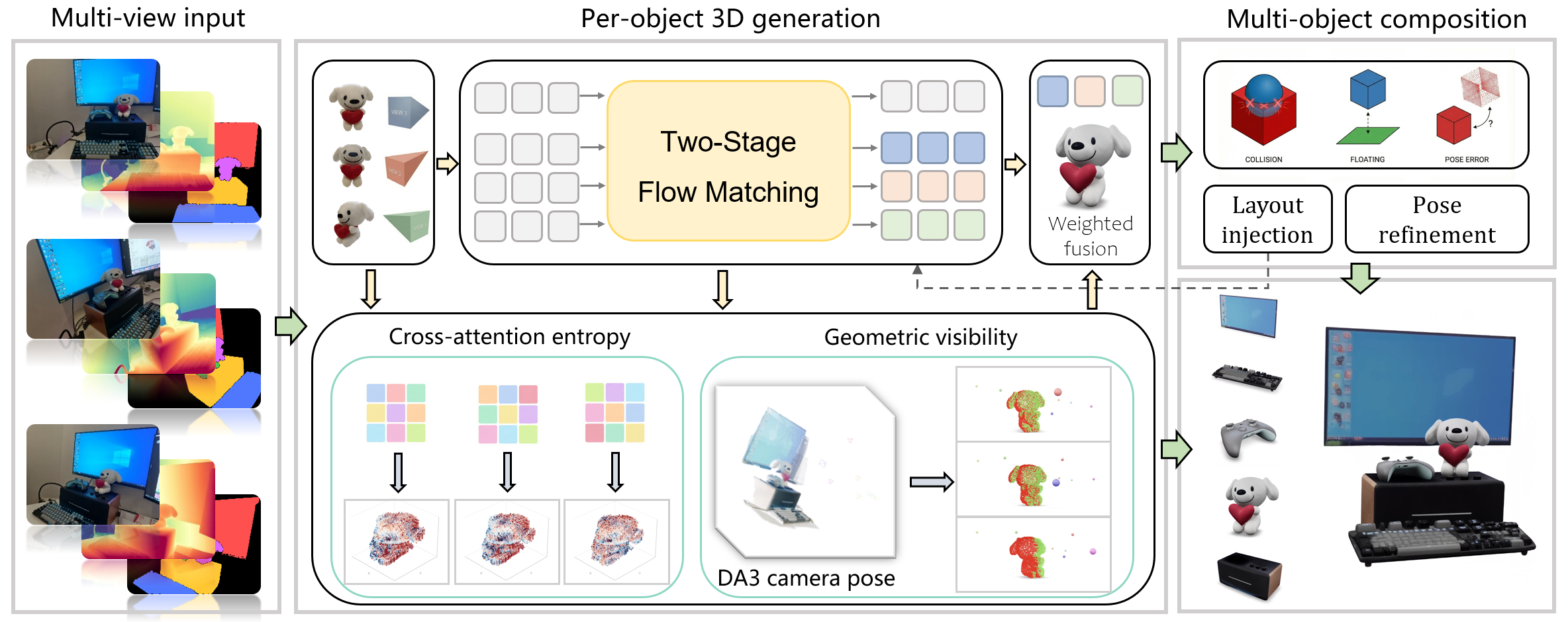}
    \caption{\textbf{Overview of MV-SAM3D.} Given multi-view images with segmentation masks and DA3-estimated pointmaps, our framework first performs per-object 3D generation by fusing flow matching velocities from each viewpoint with adaptive weighting (cross-attention entropy and geometric visibility). Multi-object composition is then achieved through layout injection during generation and post-generation pose refinement, resolving collisions, floating artifacts, and pose errors.}
    \label{fig:pipeline}
\end{figure*}

We present MV-SAM3D, a training-free framework that enables multi-view consistent and physically plausible 3D scene generation. An overview is illustrated in~\cref{fig:pipeline}. Given multi-view images with segmentation masks, we first recover metric-scale pointmaps and camera poses via multi-view 3D reconstruction (DA3~\cite{yang2024da3}). Our pipeline then performs \textbf{per-object 3D generation} with adaptive multi-view fusion (\cref{sec:adaptive_fusion}), followed by physically plausible \textbf{multi-object composition} through layout injection and post-refinement (\cref{sec:physics}). We begin with a brief review of the SAM3D pipeline, which serves as the foundation of our framework.

\subsection{Preliminaries: SAM3D}
\label{sec:preliminaries}

SAM3D generates 3D assets through a two-stage pipeline:

\noindent\textbf{Stage~1 --- Sparse Structure (SS) Generation.}
Given an input image $I$ and a segmentation mask $M$, the first stage generates a sparse voxel structure $\mathcal{V} \in \mathbb{R}^{K_1 \times d_1}$ that captures the coarse geometry of the target object. This stage uses a Multi-Modal Diffusion Transformer (MM-DiT) to jointly model the object's \emph{shape} and \emph{layout} (pose parameters $s, R, t$) through flow matching. The conditioning consists of visual features extracted from the input image and mask (both cropped and full-scene), as well as a pointmap encoding spatial correspondence.

\noindent\textbf{Stage~2 --- Structured Latent (SLAT) Generation.}
Conditioned on the sparse structure from Stage~1, the second stage generates high-resolution structured latent representations $\mathcal{Z} \in \mathbb{R}^{K_2 \times d_2}$ that encode detailed geometry and texture. This stage uses a DiT architecture with cross-attention between the latent tokens and the conditioning features from the input image.

Both stages employ \textbf{flow matching}~\cite{lipman2023flow} for generation. Given a condition $c$ derived from the input image and mask, the model learns a velocity field $v_\theta(x_t, t, c)$ that transports samples from noise $x_0 \sim \mathcal{N}(0, I)$ toward the data distribution. Generation proceeds by integrating the ODE from $t{=}0$ to $t{=}1$:
\begin{equation}
    x_{t+\Delta t} = x_t + v_\theta(x_t, t, c) \cdot \Delta t.
    \label{eq:flow_matching}
\end{equation}

\subsection{Adaptive Multi-View Fusion}
\label{sec:adaptive_fusion}

Given $N$ input images $\{I_i\}_{i=1}^N$ of an object captured from different viewpoints, with corresponding segmentation masks $\{M_i\}$ and camera poses $\{P_i\}$, our goal is to generate a single 3D asset that is consistent with all observations.

\noindent\textbf{Multi-Diffusion in 3D Latent Space.}
We extend single-view generation to the multi-view setting by adapting the Multi-Diffusion~\cite{bartal2023multidiffusion} framework to 3D structured latent space. Multiple viewpoints provide different conditioning signals $\{c_i\}_{i=1}^N$ for the \emph{same} underlying 3D latent variable $x_t$. At each flow matching step, we query the velocity field with each condition independently and fuse the predictions via a weighted combination:
\begin{equation}
    \hat{v}(x_t, t) = \sum_{i=1}^{N} w_i \cdot v_\theta(x_t, t, c_i),
    \label{eq:multi_diffusion}
\end{equation}
where $w_i$ are per-view fusion weights satisfying $\sum_i w_i = 1$. In its original formulation, Multi-Diffusion employs uniform weights $w_i = 1/N$, \ie, \textbf{naive averaging} across all viewpoints. This simple strategy already enables multi-view consistent generation in many cases and serves as our baseline for extending SAM3D to multi-view input.

\noindent\textbf{Limitations of Naive Averaging.}
However, naive averaging treats all viewpoints equally at every spatial location, regardless of whether a given 3D point is actually observed or merely hallucinated from a particular view. For a point visible from view~$j$ but occluded in view~$k$, the reliable prediction from $j$ is blended with the hallucinated one from $k$, degrading quality. This is especially problematic when views are unevenly distributed: the majority viewpoints overwhelm the minority that uniquely observe certain regions, producing structural errors and texture artifacts. As illustrated in~\cref{fig:entropy_ablation} (left column), naive averaging under a skewed view distribution can produce incorrect geometry and miss observed details entirely.

These limitations highlight the need for \emph{adaptive} fusion: rather than treating all views equally, the key question is \emph{which viewpoint's prediction is most reliable at each 3D point?} To this end, we propose two complementary weighting strategies to estimate per-point, per-view observation confidence --- one implicit, based on cross-attention patterns, and one explicit, based on geometric visibility.

\subsubsection{Attention-Entropy Weighting (Implicit Observation Confidence)}
\label{sec:entropy_weighting}

In the generation pipeline, each latent point attends to image patch tokens via cross-attention. We observe that the entropy of this attention distribution serves as a natural, implicit indicator of observation confidence. When a 3D point corresponds to a region directly captured in the input image, the model can locate the relevant image patches and attend to them with high specificity --- the attention pattern is concentrated on a small number of tokens, resulting in low entropy. Conversely, when a 3D point lies in a region \emph{not} captured by the image, no specific patch carries the relevant information; the model must instead aggregate global context from across the entire image to infer the content, producing a diffuse attention pattern with high entropy.

\begin{figure}[t]
    \centering
    \includegraphics[width=\linewidth]{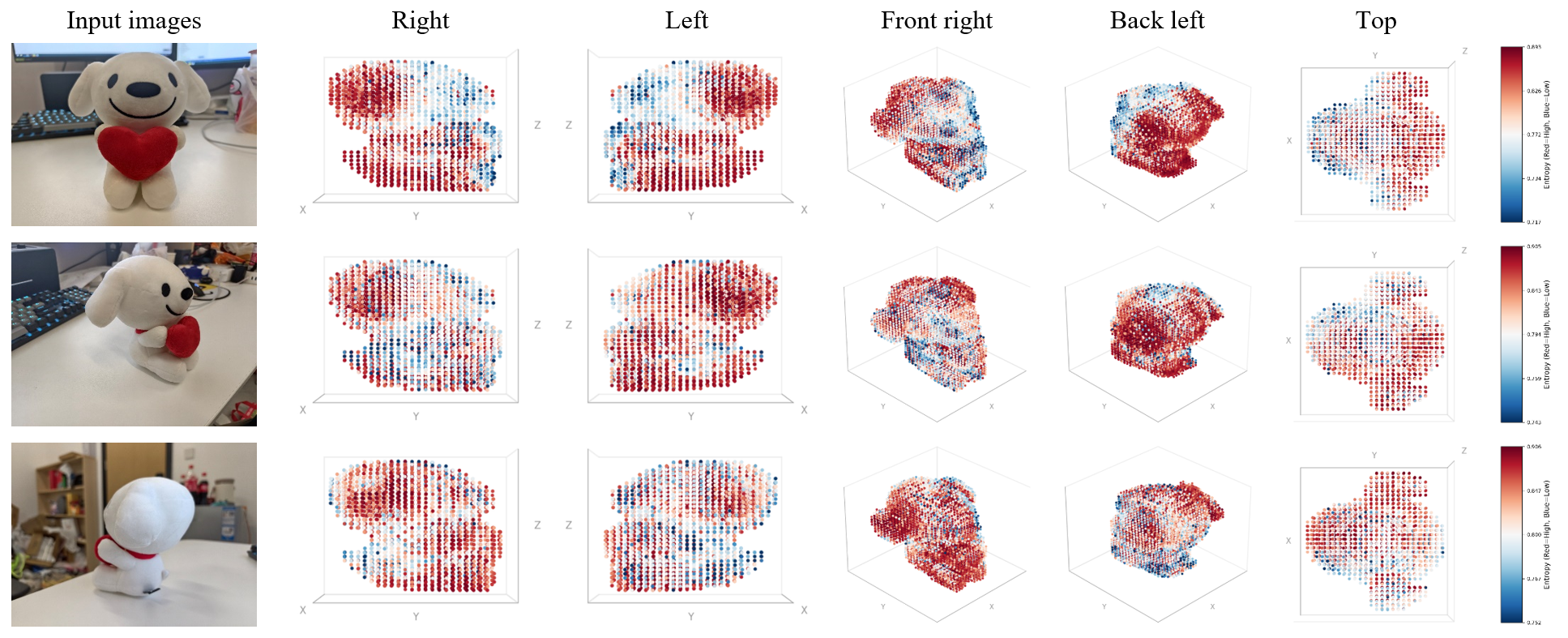}
    \caption{\textbf{Attention-entropy visualization.} For a plush toy observed from three viewpoints, we visualize the per-point cross-attention entropy. Regions directly visible from a given view exhibit low entropy (blue), while occluded regions show high entropy (red), confirming that attention entropy serves as a reliable implicit indicator of observation confidence.}
    \label{fig:entropy_vis}
\end{figure}

We visualize this in~\cref{fig:entropy_vis}: for a plush toy observed from three viewpoints, visible regions consistently exhibit low entropy while occluded regions show high entropy, confirming that attention entropy reliably reflects observation confidence.

Concretely, for each viewpoint $i$ and each latent point $l$, we extract the cross-attention weights $\mathbf{a}_{i,l} \in \mathbb{R}^{P}$ over the $P$ image patch tokens from a designated transformer layer and compute the normalized Shannon entropy:
\begin{equation}
    H_i(l) = -\frac{1}{\log P} \sum_{p=1}^{P} \hat{a}_{i,l}^{(p)} \log \hat{a}_{i,l}^{(p)},
    \label{eq:entropy}
\end{equation}
where $\hat{\mathbf{a}}_{i,l}$ is the attention distribution normalized to sum to 1 over patches. The entropy $H_i(l) \in [0, 1]$ is low when attention is concentrated (high observation confidence) and high when it is diffuse (low observation confidence).

The entropy-based fusion weight for view $i$ at latent point $l$ is:
\begin{equation}
    w_i^{\text{ent}}(l) = \frac{\exp(-\alpha \cdot H_i(l))}{\sum_{j=1}^{N} \exp(-\alpha \cdot H_j(l))},
    \label{eq:entropy_weight}
\end{equation}
where $\alpha > 0$ is a temperature parameter controlling the sharpness of the weight distribution. Higher $\alpha$ produces more polarized weights that strongly favor the most confident view.

\begin{figure}[t]
    \centering
    \includegraphics[width=\linewidth]{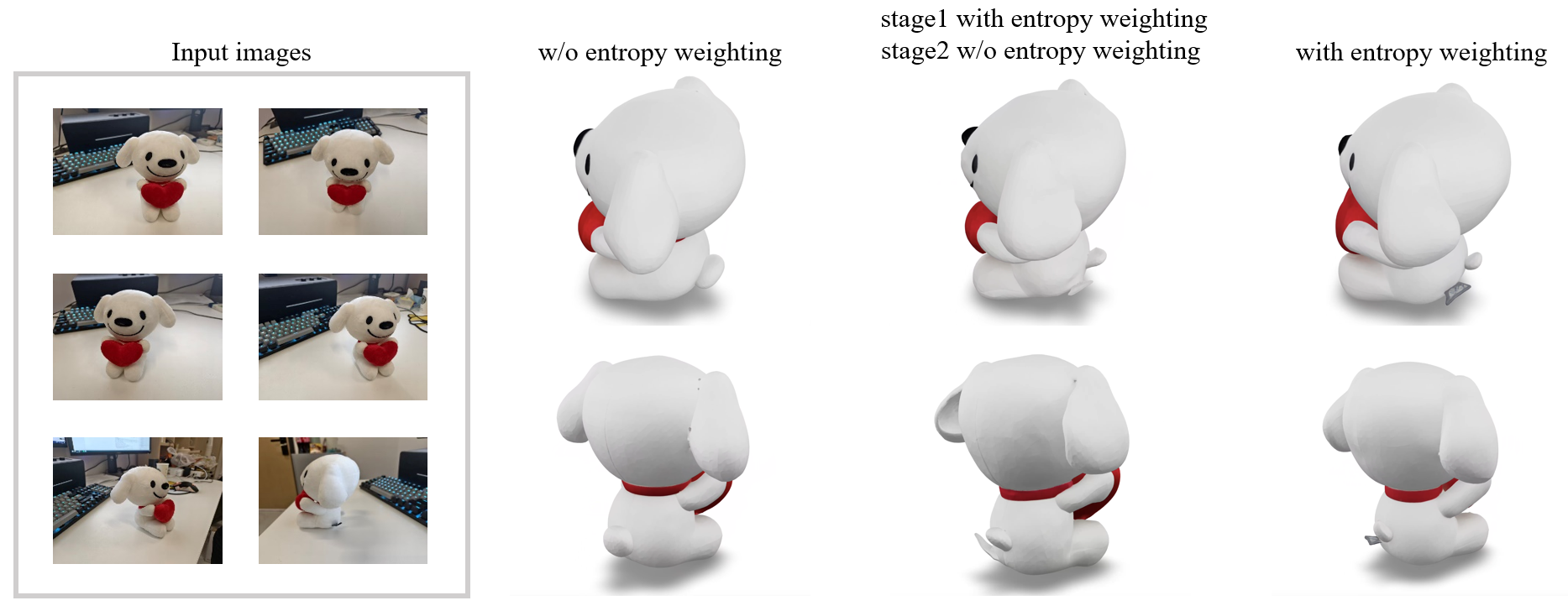}
    \caption{\textbf{Effect of entropy weighting.} A plush toy observed from 6 views (5 frontal, 1 rear capturing the tail and a black label). \textbf{Simple averaging}: tail shape is wrong and the black label is missing. \textbf{Entropy in Stage~1 only}: correct structure emerges but label texture is white. \textbf{Entropy in both stages}: both structure and texture faithfully match the observation, confirming that entropy weighting is essential in both stages.}
    \label{fig:entropy_ablation}
\end{figure}

\cref{fig:entropy_ablation} demonstrates the effectiveness of this strategy using a plush toy observed from 6 views (5 frontal, 1 rear capturing a tail and black label). With simple averaging, the frontal majority dominates: the tail shape is wrong and the label is missing. Applying entropy weighting in Stage~1 only recovers correct structure, but the label texture remains white since Stage~2 still uses uniform averaging. Enabling entropy weighting in \emph{both stages} produces correct structure \emph{and} correct black texture, confirming that confidence weighting is essential in both generation stages.

\subsubsection{Visibility Weighting (Explicit Observation Confidence)}
\label{sec:visibility_weighting}

While attention-entropy weighting is generally effective, it relies on the model's learned implicit matching between 3D points and image patches. This implicit process can be disrupted in certain cases --- for instance, objects with symmetric or repetitive structures may cause the model to confidently attend to a visually similar but geometrically incorrect region, producing low entropy despite the point not being truly observed.

\begin{figure}[t]
    \centering
    \includegraphics[width=\linewidth]{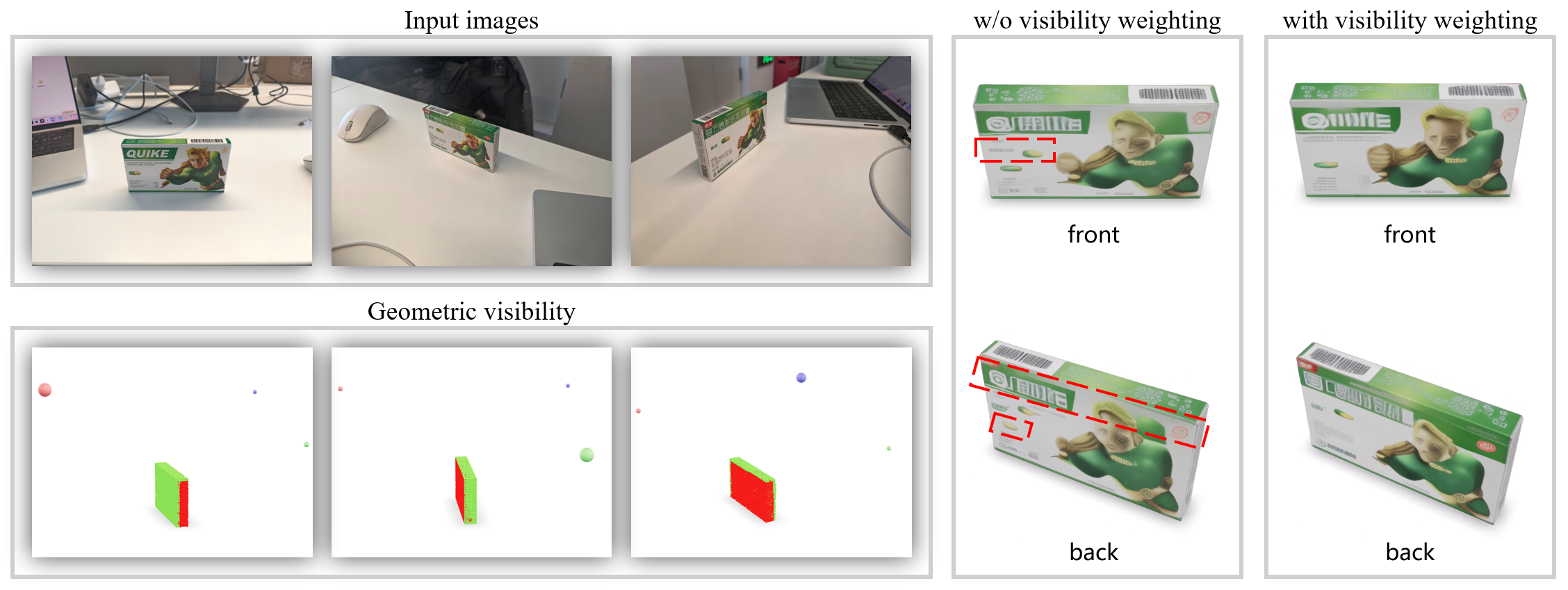}
    \caption{\textbf{Effect of visibility weighting.} A medicine box with distinct front/back textures. \textbf{Entropy weighting only}: front and back textures are mixed due to the symmetric structure confusing implicit matching. \textbf{Entropy + visibility weighting}: front and back appearances are correctly separated, with each face faithfully reflecting the observed texture.}
    \label{fig:visibility_ablation}
\end{figure}

We illustrate this failure mode in~\cref{fig:visibility_ablation} with a medicine box whose front and back faces carry distinct but structurally similar textures. With entropy weighting alone, the model confuses front and back --- both sides receive mixed textures, with features from one face bleeding onto the other. The implicit attention-based matching cannot reliably distinguish between the two visually similar but geometrically opposite surfaces.

To complement the implicit approach, we introduce an explicit strategy based on geometric visibility that becomes available once the coarse 3D structure is generated. Since our layout-aware generation recovers each object's pose relative to the primary input viewpoint, we can combine this with the known inter-camera relative poses to transform the voxel structure into every camera's coordinate frame and determine which latent points are geometrically visible from each viewpoint --- a purely geometric computation independent of the model's internal behavior, readily applicable to the subsequent generation process.

Given the sparse voxel structure $\mathcal{V}$ from Stage~1 and camera poses $\{P_i\}$, we perform DDA ray tracing to compute a binary visibility matrix $V \in \{0, 1\}^{N \times K}$, where $V_{i,l} = 1$ if latent point $l$ is geometrically visible from camera $i$ and $V_{i,l} = 0$ if it is self-occluded. The visibility-based weight is:
\begin{equation}
    w_i^{\text{vis}}(l) = \frac{\exp(\beta \cdot V_{i,l})}{\sum_{j=1}^{N} \exp(\beta \cdot V_{j,l})},
    \label{eq:visibility_weight}
\end{equation}
where $\beta > 0$ controls the penalty for occlusion. For large $\beta$, the weight of an occluded viewpoint is effectively suppressed to zero.

As shown in~\cref{fig:visibility_ablation}, adding visibility weighting correctly separates front and back appearances: the front displays only its own green banner in the upper-left, while the back shows its distinct full-width banner --- faithfully matching the real observations.

Both strategies address the same question --- which viewpoint to trust at each location --- but through different signals. Attention entropy provides a soft, implicit estimate from the model's behavior, broadly applicable but susceptible to visual ambiguities. Visibility provides a hard, explicit estimate grounded in 3D geometry, immune to appearance confusion. Their combination yields a robust confidence estimate:
\begin{equation}
    w_i(l) = (1 - \gamma) \cdot w_i^{\text{ent}}(l) + \gamma \cdot w_i^{\text{vis}}(l),
    \label{eq:combined_weight}
\end{equation}
where $\gamma \in [0, 1]$ balances the two signals.

\subsection{Physics-Aware Pose Optimization}
\label{sec:physics}

The adaptive fusion pipeline described above handles per-object generation, but independently estimated poses may still cause inter-object collisions or physically implausible arrangements. We address this through two complementary mechanisms.

\subsubsection{Layout Injection}
\label{sec:layout_injection}

To proactively avoid physical conflicts, we inject gradient-based guidance into the structure generation process where geometry and layout are jointly determined, analogous to classifier guidance~\cite{dhariwal2021diffusion} but with physical plausibility as the objective. During the later portion of the flow matching trajectory, we periodically decode the current latent to obtain approximate voxel occupancy $\tilde{\mathcal{V}}^{(k)}_t$ and evaluate a physics loss against other objects that share physical interactions (\eg, contact):
\begin{equation}
    \mathcal{L}_{\text{phys}} = \mathcal{L}_{\text{collision}}\!\left(\tilde{\mathcal{V}}^{(k)}_t, \{\mathcal{V}^{(j)}\}_{j \neq k}\right) + \lambda_{\text{contact}} \mathcal{L}_{\text{contact}}\!\left(\tilde{\mathcal{V}}^{(k)}_t, \{\mathcal{V}^{(j)}\}_{j \neq k}\right),
    \label{eq:physics_loss}
\end{equation}
where $\mathcal{L}_{\text{collision}}$ penalizes volumetric overlap and $\mathcal{L}_{\text{contact}}$ encourages proximate surfaces to maintain physical contact. The gradient of this loss is then used to steer the flow matching trajectory:
\begin{equation}
    x_{t+\Delta t} = x_t + v_\theta(x_t, t, c) \cdot \Delta t - \eta \nabla_{x_t} \mathcal{L}_{\text{phys}},
    \label{eq:guided_flow}
\end{equation}
where $\eta$ controls the guidance strength. This proactively avoids physical conflicts before they are baked into the final output.

\subsubsection{Post-Generation Pose Refinement}
\label{sec:post_refinement}

While layout injection reduces physical conflicts during generation, the discrete nature of latent-space guidance may leave residual violations. After all objects are generated, we perform a global pose refinement directly on the output meshes. For each object $k$, we optimize its similarity transform $\theta_k = (s_k, R_k, t_k)$ while keeping mesh geometry fixed, minimizing:
\begin{equation}
    \mathcal{L}_{\text{total}} = \sum_k \mathcal{L}_{\text{align}}^{(k)} + \lambda_{\text{col}} \sum_{j \neq k} \mathcal{L}_{\text{col}}^{(j,k)} + \lambda_{\text{con}} \sum_{j \neq k} \mathcal{L}_{\text{con}}^{(j,k)} + \lambda_{\text{reg}} \mathcal{L}_{\text{reg}}.
    \label{eq:total_loss}
\end{equation}
The alignment term $\mathcal{L}_{\text{align}}$ measures Chamfer Distance between each transformed mesh and its corresponding point cloud from metric-scale depth maps, while $\mathcal{L}_{\text{col}}$ and $\mathcal{L}_{\text{con}}$ penalize inter-object penetration and encourage surface contact, respectively. $\mathcal{L}_{\text{reg}}$ regularizes against large deviations from the initial pose estimates.

%% file: secs/4_experiments.tex
\section{Experiments}
\label{sec:experiments}

\subsection{Experimental Setup}
\label{sec:setup}

All experiments are conducted on a single NVIDIA A100 (80GB) GPU. We follow SAM3D's default configuration for the base generation pipeline. For the components introduced in MV-SAM3D, we set the entropy weighting temperature $\alpha = 30$, visibility weighting parameter $\beta = 30$, and mixing coefficient $\gamma = 0.5$. For layout injection, guidance begins at step $T_s = 15$ (out of 25 total steps) with guidance interval $\Delta_g = 3$ and strength $\eta = 0.1$, with $\lambda_{\text{contact}} = 0.5$. For post-generation pose refinement, we optimize with Adam for 100 iterations (learning rate $0.01$, collision weight $\lambda_{\text{col}} = 200$, contact weight $\lambda_{\text{con}} = 50$, proximity threshold $\tau = 0.05$).

We evaluate on two datasets. The first is the \textbf{GSO benchmark}~\cite{downs2022gso}, consisting of Google Scanned Objects with ground-truth 3D models, from which we render multi-view images at known camera poses. The second is \textbf{MV-SAM3D-Scenes}, a self-captured multi-object dataset containing real-world scenes with 2 to 8 objects in diverse configurations, each captured from 15 viewpoints. We compare against representative baselines including SAM3D~\cite{sam3d}, TRELLIS~\cite{trellis}, DreamGaussian~\cite{tang2024dreamgaussian}, SyncDreamer~\cite{liu2024syncdreamer}, and EscherNet~\cite{kong2024eschernet}. For single-object evaluation on GSO, we adopt the evaluation protocol and metrics from EscherNet~\cite{kong2024eschernet} and report baseline results from the respective publications alongside ours for fair comparison.

\subsection{Multi-View Single-Object Generation}
\label{sec:exp_single}

The qualitative analyses in~\cref{sec:adaptive_fusion} have already illustrated the role of each weighting strategy through targeted examples:~\cref{fig:entropy_vis} visualizes the correlation between attention entropy and observation confidence,~\cref{fig:entropy_ablation} confirms that entropy weighting in both stages is essential for faithful reconstruction, and~\cref{fig:visibility_ablation} demonstrates that visibility weighting further resolves ambiguities arising from symmetric structures. Here we present systematic evaluation on the GSO benchmark.

\cref{tab:gso_cd,tab:gso_nvs} report quantitative results. Multi-view methods clearly outperform all single-view baselines, confirming that additional observations substantially reduce hallucination. Among multi-view approaches, MV-SAM3D consistently outperforms EscherNet across most metrics, with particularly favorable margins given fewer input views. TRELLIS with naive Multi-Diffusion fusion consistently lags behind, demonstrating that confidence-aware weighting is essential for effective multi-view fusion. Beyond metric performance, MV-SAM3D additionally provides layout-aware generation --- preserving each object's spatial arrangement within the scene --- a capability that pure reconstruction methods like EscherNet lack.

\begin{table}[t]
    \centering
    \scriptsize
    \renewcommand{\arraystretch}{0.95}
    \setlength{\tabcolsep}{2pt}
    \begin{minipage}[t]{0.40\linewidth}
        \centering
        \caption{CD ($\times 10^{-3}$) on GSO-30.}
        \label{tab:gso_cd}
        \begin{tabular}{lcc}
            \toprule
            Method & \#V & CD $\downarrow$ \\
            \midrule
            DreamGauss.~\cite{tang2024dreamgaussian} & 1 & 60.5 \\
            DG-XL~\cite{tang2024dreamgaussian} & 1 & 45.9 \\
            SyncDreamer~\cite{liu2024syncdreamer} & 1 & 40.0 \\
            SAM3D~\cite{sam3d} & 1 & 42.0 \\
            TRELLIS~\cite{trellis} & 1 & 44.5 \\
            \midrule
            EscherNet~\cite{kong2024eschernet} & 2 & \underline{21.5} \\
            TRELLIS+M.D. & 2 & 24.8 \\
            Ours & 2 & \textbf{20.2} \\
            \cmidrule{1-3}
            EscherNet~\cite{kong2024eschernet} & 5 & \underline{17.5} \\
            TRELLIS+M.D. & 5 & 21.3 \\
            Ours & 5 & \textbf{17.3} \\
            \bottomrule
        \end{tabular}
    \end{minipage}%
    \hfill
    \begin{minipage}[t]{0.58\linewidth}
        \centering
        \caption{Novel view synthesis on GSO-30.}
        \label{tab:gso_nvs}
        \begin{tabular}{lcccc}
            \toprule
            Method & \#V & PSNR $\uparrow$ & SSIM $\uparrow$ & LPIPS $\downarrow$ \\
            \midrule
            RealFusion~\cite{melaskyriazi2023realfusion} & 1 & 12.76 & 0.758 & 0.382 \\
            Zero123~\cite{liu2023zero123} & 1 & 18.51 & 0.856 & 0.127 \\
            Zero123-XL~\cite{deitke2023objaverse} & 1 & 18.93 & 0.856 & 0.124 \\
            SAM3D~\cite{sam3d} & 1 & 19.82 & 0.872 & 0.108 \\
            TRELLIS~\cite{trellis} & 1 & 19.15 & 0.862 & 0.118 \\
            \midrule
            EscherNet~\cite{kong2024eschernet} & 2 & \underline{22.91} & \underline{0.908} & \underline{0.064} \\
            TRELLIS+M.D. & 2 & 21.35 & 0.889 & 0.081 \\
            Ours & 2 & \textbf{23.45} & \textbf{0.912} & \textbf{0.059} \\
            \cmidrule{1-5}
            EscherNet~\cite{kong2024eschernet} & 5 & \underline{25.09} & \textbf{0.927} & \underline{0.043} \\
            TRELLIS+M.D. & 5 & 23.15 & 0.908 & 0.060 \\
            Ours & 5 & \textbf{25.22} & \underline{0.925} & \textbf{0.042} \\
            \bottomrule
        \end{tabular}
    \end{minipage}
\end{table}

\begin{figure*}[t]
    \centering
    \includegraphics[width=\linewidth]{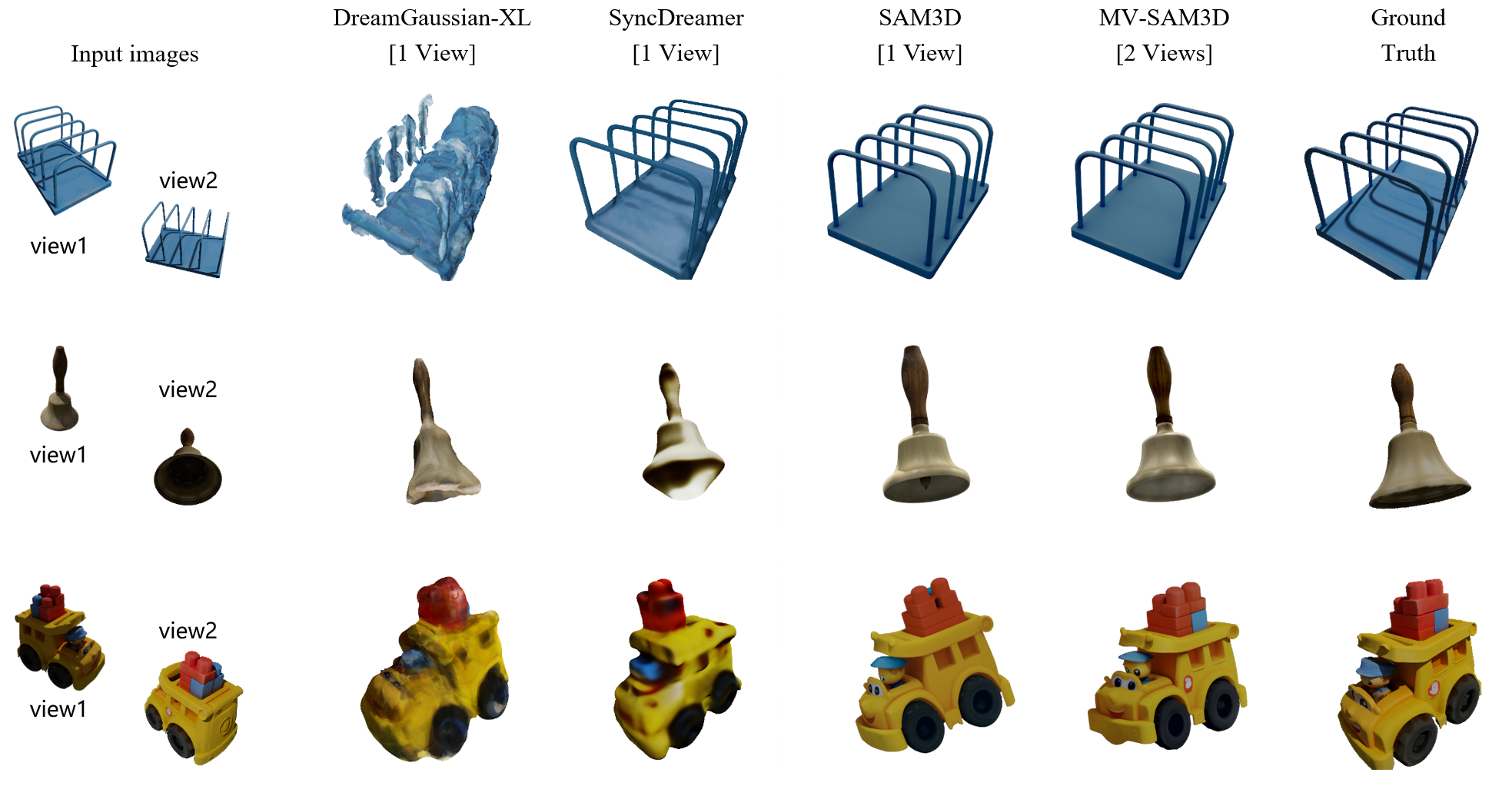}
    \caption{\textbf{Qualitative comparison with single-view methods on GSO.} Single-view methods produce plausible front-facing geometry but hallucinate textures on unobserved sides. MV-SAM3D leverages multi-view observations to produce reconstructions faithful to the true object appearance.}
    \label{fig:single_view_comp}
\end{figure*}

\begin{figure*}[t]
    \centering
    \includegraphics[width=\linewidth]{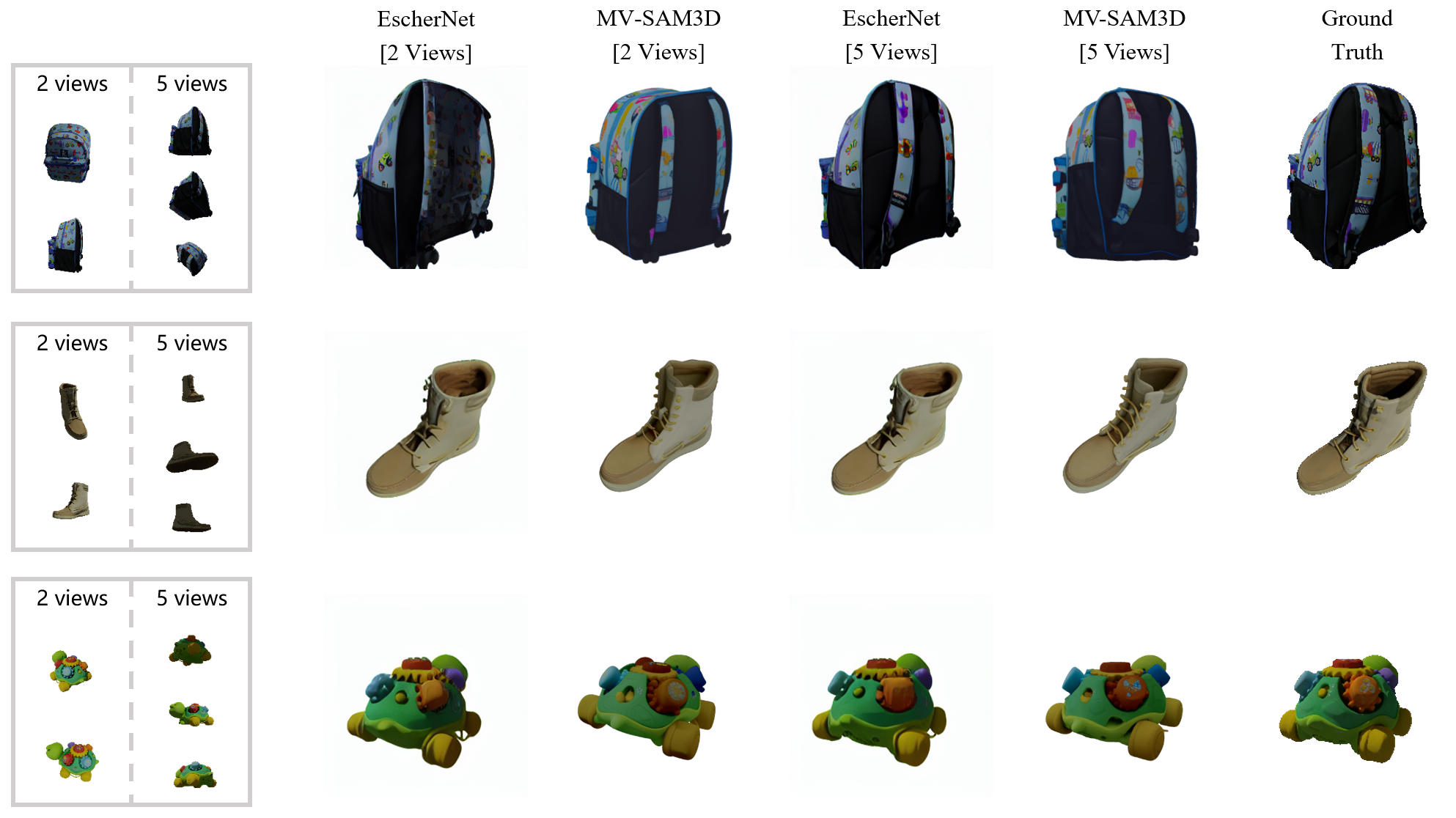}
    \caption{\textbf{Qualitative comparison with EscherNet on GSO.} Both methods produce reasonable reconstructions from 2 and 5 input views. Our results exhibit slightly more faithful texture details in certain regions. Quantitative comparison is provided in~\cref{tab:gso_cd,tab:gso_nvs}.}
    \label{fig:eschernet_comp}
\end{figure*}

\cref{fig:single_view_comp} provides qualitative comparisons with single-view methods on GSO objects. Single-view approaches produce plausible front-facing geometry but generate hallucinated texture on unobserved sides, while MV-SAM3D produces reconstructions consistent with the true object appearance from all observed angles. \cref{fig:eschernet_comp} further compares with EscherNet using 2 and 5 input views. Both methods produce reasonable reconstructions; our results exhibit more faithful texture details in certain regions.

\subsection{Multi-Object Scene Composition}
\label{sec:exp_multi}

We evaluate the complete MV-SAM3D pipeline on multi-object scenes from our MV-SAM3D-Scenes dataset.~\cref{fig:teaser} (left) provides an overview of a representative scene-level reconstruction, where each generated 3D object is overlaid onto the scene point cloud, demonstrating the overall quality of our multi-view, multi-object generation.

\begin{figure*}[t]
    \centering
    \includegraphics[width=\linewidth]{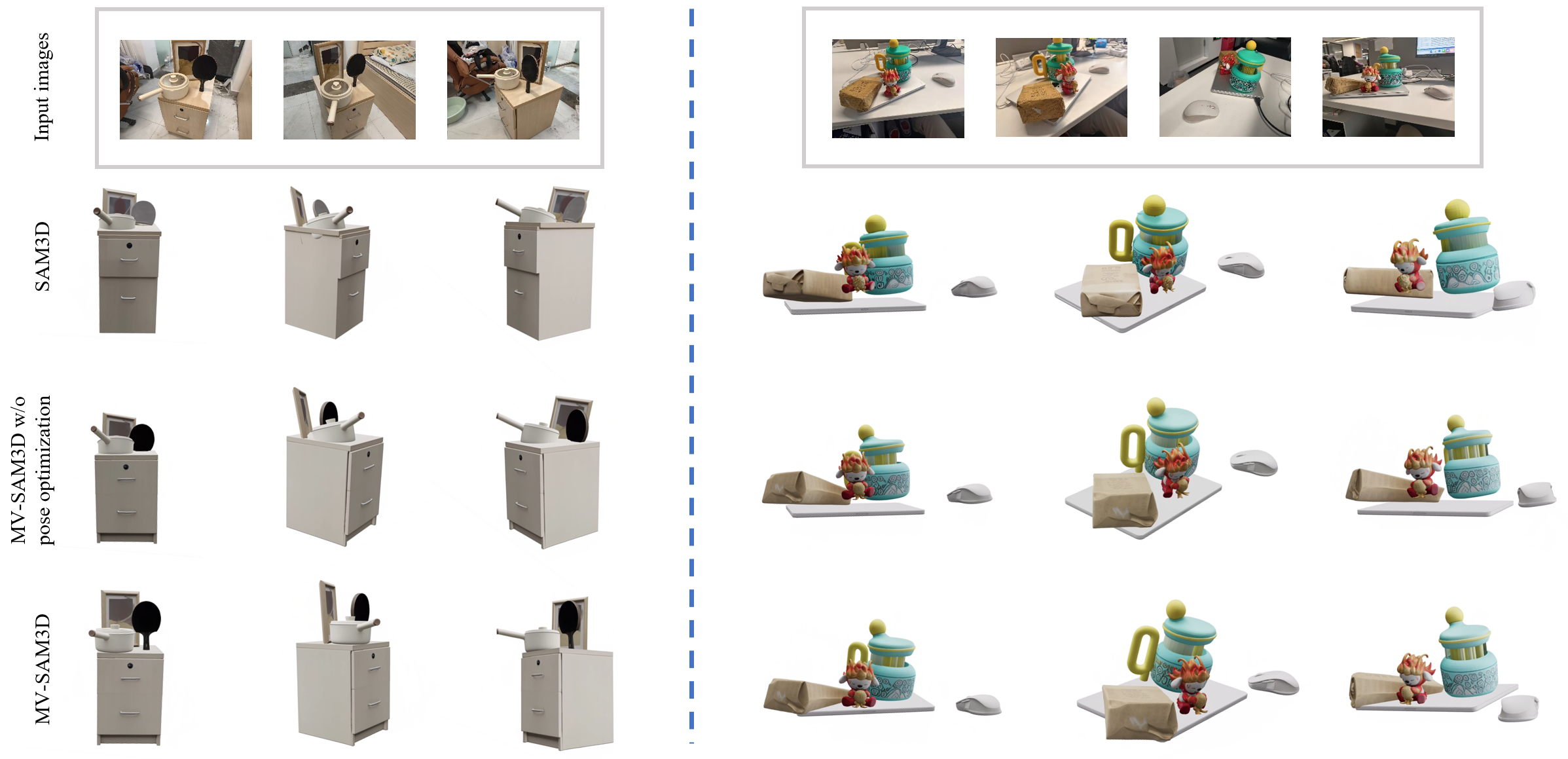}
    \caption{\textbf{Multi-object scene composition.} Comparison of SAM3D, MV-SAM3D without pose optimization, and full MV-SAM3D. SAM3D produces geometric errors and layout artifacts (collisions, floating). Multi-view fusion improves per-object geometry but layout issues persist. Our full pipeline achieves both faithful geometry and physically plausible object arrangements.}
    \label{fig:multi_object_comp}
\end{figure*}

\cref{fig:multi_object_comp} compares three configurations. SAM3D (single-view) produces noticeable geometric errors and frequent collisions/floating artifacts. Adding multi-view fusion (MV-SAM3D w/o pose optimization) improves per-object geometry through more reliable observations, but inter-object layout issues persist since poses are not yet refined. Our full pipeline achieves both faithful geometry and physically plausible layouts: surfaces in contact are touching and collisions are resolved (see also~\cref{fig:teaser}). This illustrates the complementary nature of our contributions: multi-view fusion improves \emph{per-object quality}, while physics-aware optimization ensures \emph{inter-object physical plausibility}.

Quantitative results on MV-SAM3D-Scenes are summarized in~\cref{tab:multi_object}. Multi-view fusion alone (MV-SAM3D w/o Opt.) already yields moderate improvements over single-view SAM3D in both per-object quality and spatial alignment, demonstrating that multi-view observations provide more reliable geometry and initial pose estimates. The most substantial gains come from physics-aware optimization, which dramatically improves all spatial alignment metrics, confirming its essential role in resolving inter-object layout violations.

\begin{table}[t]
    \centering
    \caption{Multi-object scene evaluation on MV-SAM3D-Scenes.}
    \label{tab:multi_object}
    \resizebox{\linewidth}{!}{%
    \begin{tabular}{lcccccc}
        \toprule
        Method & PSNR$\uparrow$ & Depth MAE$\downarrow$ & Depth RMSE$\downarrow$ & Acc@5cm$\uparrow$ & RelAcc@5\%$\uparrow$ & Cov.$\uparrow$ \\
        \midrule
        SAM3D & 19.52 & 0.079 & 0.107 & 55.66 & 44.07 & 61.20 \\
        MV-SAM3D w/o Opt. & 19.95 & 0.068 & 0.095 & 60.23 & 49.52 & 66.48 \\
        MV-SAM3D (Full) & \textbf{21.83} & \textbf{0.025} & \textbf{0.055} & \textbf{91.66} & \textbf{91.13} & \textbf{97.22} \\
        \bottomrule
    \end{tabular}%
    }
\end{table}

\subsection{Ablation Studies}
\label{sec:ablation}

\noindent\textbf{Adaptive weighting strategies.}
\cref{tab:ablation_weighting} ablates each component of our multi-view fusion on MV-SAM3D-Scenes with 5 input views. Adding entropy weighting in Stage~1 primarily improves structural accuracy, as Stage~1 governs the coarse geometry that determines spatial alignment. Extending it to Stage~2 further recovers texture fidelity, consistent with the qualitative progression in~\cref{fig:entropy_ablation}. Incorporating visibility weighting in Stage~2 yields additional appearance gains, particularly for objects with symmetric structures where the implicit entropy signal can be unreliable (cf.~\cref{fig:visibility_ablation}).

\begin{table}[t]
    \centering
    \footnotesize
    \setlength{\tabcolsep}{3pt}
    \begin{minipage}[t]{0.52\linewidth}
        \centering
        \caption{Ablation on adaptive weighting strategies (MV-SAM3D-Scenes, 5 views).}
        \label{tab:ablation_weighting}
        \begin{tabular}{llcc}
            \toprule
            Stage 1 & Stage 2 & PSNR $\uparrow$ & Acc@5cm $\uparrow$ \\
            \midrule
            Avg & Avg & 20.35 & 88.42 \\
            Entropy & Avg & 20.52 & 91.48 \\
            Entropy & Entropy & 21.35 & 91.60 \\
            Entropy & Ent.+Vis. & \textbf{21.83} & \textbf{91.66} \\
            \bottomrule
        \end{tabular}
    \end{minipage}%
    \hfill
    \begin{minipage}[t]{0.44\linewidth}
        \centering
        \caption{Effect of number of views (MV-SAM3D-Scenes).}
        \label{tab:ablation_views}
        \begin{tabular}{ccc}
            \toprule
            \#Views & PSNR $\uparrow$ & Acc@5cm $\uparrow$ \\
            \midrule
            1 & 19.52 & 85.20 \\
            2 & 21.08 & 90.45 \\
            3 & 21.52 & 91.18 \\
            5 & \textbf{21.83} & \textbf{91.66} \\
            \bottomrule
        \end{tabular}
    \end{minipage}
\end{table}

\noindent\textbf{Number of input views.}
\cref{tab:ablation_views} shows that more input views consistently improve generation quality, with the most pronounced gain from one to two views --- where the second viewpoint reveals the largest amount of previously unobserved surface. Further views yield diminishing returns, as the object is already well covered and newly added observations carry increasingly redundant information.

\noindent\textbf{Physics-aware pose optimization.}
\cref{tab:ablation_physics} ablates layout injection and post-refinement on MV-SAM3D-Scenes. Layout injection nearly eliminates inter-object collisions during generation and also improves spatial alignment metrics, providing a physically plausible initial configuration. Post-generation refinement further boosts alignment precision and coverage by directly optimizing object poses on the output meshes. Combining both achieves the best overall performance, confirming their complementary roles.

\begin{table}[t]
    \centering
    \footnotesize
    \renewcommand{\arraystretch}{0.95}
    \setlength{\tabcolsep}{3pt}
    \caption{Ablation on physics-aware pose optimization (MV-SAM3D-Scenes).}
    \label{tab:ablation_physics}
    \begin{tabular}{cccccc}
        \toprule
        Layout Inj. & Post Ref. & Acc@5cm $\uparrow$ & RelAcc@5\% $\uparrow$ & Cov. $\uparrow$ & Col. Rate $\downarrow$ \\
        \midrule
        \ding{55} & \ding{55} & 60.23 & 49.52 & 66.48 & 15.32 \\
        \ding{51} & \ding{55} & 71.45 & 60.18 & 75.25 & 1.67 \\
        \ding{55} & \ding{51} & 87.82 & 85.60 & 94.18 & \textbf{0.00} \\
        \ding{51} & \ding{51} & \textbf{91.66} & \textbf{91.13} & \textbf{97.22} & \textbf{0.00} \\
        \bottomrule
    \end{tabular}
\end{table}

\section{Conclusion}
\label{sec:conclusion}

We have presented MV-SAM3D, a training-free framework that extends layout-aware 3D generation to multi-view input with physically plausible multi-object composition. Through confidence-aware fusion --- driven by attention-entropy and geometric visibility weighting --- and physics-aware pose optimization, our approach faithfully leverages complementary observations while resolving inter-object collisions and enforcing surface contact. Experiments confirm consistent improvements in both reconstruction fidelity and layout plausibility.